\def\year{2019}\relax
\documentclass[letterpaper]{article} 
\usepackage{aaai19}  
\usepackage{times}  
\usepackage{helvet}  
\usepackage{courier}  
\usepackage{url}  
\usepackage{graphicx}  
\usepackage{amssymb}
\usepackage{amsmath}
\usepackage{amsfonts}
\usepackage{amsthm}
\usepackage{comment}
\usepackage{thmtools}
\usepackage{thm-restate}
\usepackage{mdframed}
\usepackage{booktabs}

\newcommand{\eps}{\varepsilon}
\newcommand{\NP}{\textsc{NP}}

\usepackage{xcolor}
\usepackage{colortbl}
\colorlet{tableheadcolor}{gray!25} 
\colorlet{tablerowcolor}{gray!20} 
\newcommand{\rowcol}{\rowcolor{tablerowcolor}} %

\begin{document}
%
\title{Fine-grained Search Space Classification for\\Hard Enumeration Variants of Subset Problems}

\author{
Juho Lauri \and Sourav Dutta\\
Nokia Bell Labs, Ireland\\
\{juho.lauri, sourav.dutta\}@nokia-bell-labs.com
}

\maketitle
\begin{abstract}
We propose a simple, powerful, and flexible machine learning framework for (i) reducing the search space of computationally difficult enumeration variants of subset problems and (ii) augmenting existing state-of-the-art solvers with informative cues arising from the input distribution.
We instantiate our framework for the problem of listing \emph{all} maximum cliques in a graph, a central problem in network analysis, data mining, and computational biology.
We demonstrate the practicality of our approach on real-world networks with millions of vertices and edges by not only retaining all optimal solutions, but also aggressively pruning the input instance size resulting in several fold speedups of state-of-the-art algorithms.
Finally, we explore the limits of scalability and robustness of our proposed framework, suggesting that supervised learning is viable for tackling $\NP$-hard problems in practice.
\end{abstract}

\section{Introduction}

Computationally challenging (i.e., $\NP$-hard) problems are ubiquitous and arise naturally in several domains like scheduling, planning, and design and analysis of networks. 
In particular, several such important problems are \emph{subset problems}: given an input, what is the largest/smallest subset with a particular property?
One of the most central of such problems is the \emph{maximum clique problem}.
In this problem, we are given an undirected graph $G$, and asked for the largest subset of vertices that are pairwise adjacent, i.e., form a \emph{clique}.
An even harder and more general variant of a subset problem is its \emph{enumeration} variant, in which the goal is to list \emph{all} optimal solutions.
This variant of the maximum clique problem is known as the \emph{maximum clique enumeration} (MCE) problem. Note that we consider the problem of listing of all \emph{maximum} cliques and not \emph{maximal} cliques.

The MCE problem has numerous applications such as social network analysis~\cite{soc}, study of structures in behavioral networks~\cite{beha}, statistical analysis of financial networks~\cite{finan}, and clustering in citation and dynamic networks~\cite{dynamic}. 
Moreover, MCE is also closely tied to various applications in computational 
biology~\cite{bio,Eblen2012,Yeger2004}, including the detection of protein-protein interaction complex, clustering protein sequences, searching for common cis-regulatory elements~\cite{protein}, and others~\cite{mce}.

Unfortunately, in the worst case, $\NP$-hard subset problems like maximum clique are inherently complex even for small datasets and lack scalable algorithms. 
To make matters worse, high-throughput data typically creates extremely large graphs in e.g., data mining and computational biology~\cite{Eblen2012}.
State-of-the-art solvers employ various heuristics that are effective when the input graph has certain structural properties, but the detection of such structure can be costly in practice.
In addition, the solvers are general purpose, and are unable to exploit information concerning the input distribution.

We argue that practically relevant problem instances typically come from the same distribution due to e.g., human nature or laws of physics governing the process the input models.
For instance, it is plausible that humans connect in similar ways on a social network, or that different human lungs respond similarly to cigarette smoke.
Can we augment existing state-of-the-art solvers to cheaply and automatically leverage such information about the input distribution, which we cannot even describe?
In particular, \textit{can machine learning be used to cheaply detect and exploit structure in practically relevant instances of $\NP$-hard problems that come from the same distribution?}

To the best of our knowledge, our work is the first attempt at using machine learning to reduce the search space of a computationally difficult enumeration problem.

\subsection{Related work}

\paragraph{Indirect approaches}
``Indirect approaches'' are unable to solve hard problem instances themselves, but instead they augment existing solvers, for problems such as Boolean satisfiability and mixed-integer programming.
In particular, such methods have been applied in terms of restart strategies~\cite{Gomes1998,Gomes2000}, branching heuristics~\cite{Liang2016,He2014,Khalil2016}, parameter tuning~\cite{Hutter2009}, and algorithm portfolios~\cite{Fitzgerald2015,Loreggia2016}.
In addition, applications of machine learning to \emph{discover} algorithms have also been successful~\cite{Khalil2017}, but only for small graph sizes (up to 1200 vertices).

\paragraph{Direct approaches}
By a ``direct approach'' we refer to an approach that can solve a hard problem by itself.
Various researchers proposed approaches for solving TSP~\cite{Hopfield1985,Fort1988,Durbin1987}.
However, these methods can not compete with direct algorithmic methods for TSP like the Concorde solver~\cite{Applegate2006}.
Similarly, there has been interest in studying the power of neural networks for solving hard problems~\cite{Bruck1988,Yao1992}.

Recently, there has been interest in bringing advancements from deep learning and game-playing to combinatorial optimization~\cite{Vinyals2015,Bello2016,Nowak2017}.
These approaches apply for small graph sizes (up to only 200 vertices), whereas direct methods handle instances with several tens of thousands of vertices~\cite{Applegate2009}.

Another approach for solving hard problems is via supervised learning, by treating a classifier as a decision oracle~\cite{Devlin2008,Xu2018}.
The challenge in making use of such an oracle is that one is rarely satisfied by merely an affirmative answer, but demands a witness as well.

\paragraph{The MCE problem}
The maximum clique problem is heavily studied and is well-known to be $\NP$-complete with other strong hardness results known as well~\cite{Zuckerman2006,Chen2006}.
The MCE problem is at least as hard as the maximum clique problem, since its solution includes all maximum cliques.

Unlike \emph{maximal} clique enumeration, MCE has received significantly less attention. 
Since any algorithm that enumerates all maximal cliques also enumerates all maximum cliques, it is natural to approach MCE by adapting the existing maximal clique enumeration algorithms~\cite{mce}. 
However, this approach quickly becomes infeasible for large dense graphs.

Existing approaches for maximal clique enumeration problem can be broadly classified into two strategies: iterative enumeration~\cite{cliq2,bit} and backtracking~\cite{akk,bron,sur,max2}, with other approaches as well~\cite{gamemce,tri,sparse,Cheng2011,lim,para,mapr}; for a comprehensive discussion on the related literature see~\cite{sur,mce}.

\subsection{Our contribution}
Our contribution overcomes the following challenges:
\begin{enumerate}
\item \textit{Whole-instance classification:} a classifier can be used as a decision oracle for a decision problem.
More precisely, one can apply the notion of self-reducibility to iteratively close-in on a solution.\footnote{For instance, given a graph $G$, imagine an oracle which answer positively iff $G$ has a $k$-clique. If the oracle answers negatively on input $G \setminus \{v\}$ for $v \in V(G)$, we have learned that $v$ is contained in a $k$-clique.}
However, it is challenging to make \emph{practical} use of such an oracle.

\item \textit{Cost of optimal labels:}
\cite[Section~4]{Bello2016} argue that learning from examples is undesirable for $\NP$-hard problems since ``getting high-quality labeled data is expensive and may be infeasible for new problem statements''.
We show, however, that if the labeled data points are representative enough of the input distribution, we can mitigate the need for labeling costly data points due to the generalizability of our classifier.
\end{enumerate}
In summary, our major contributions are as follows.
\begin{itemize}
\item A novel machine learning framework for reducing the search space of computationally hard enumeration variants of subset problems with instances drawn from the same distribution.

\item Specifically, we instantiate our framework for listing \emph{all} maximum cliques in a graph by applying computationally cheap graph-theoretic and statistical features for search space pruning via fine-grained classification of vertices.

\item We show that our framework retains \emph{all optimal solutions} on large, real-world networks with millions of vertices and edges, with significant (typically over 90 \%) reductions in the instance size, resulting in several fold speedups of state-of-the-art algorithms for the problem.

\item We explain the high accuracy of our framework by exploring its limits in terms cheap trainability, scalability, and robustness on Erd\H{o}s-R\'{e}nyi random graphs.
\end{itemize}

\section{Proposed framework}
\label{sec:framework}

We instantiate our framework for the MCE problem, but stress that the approach works in principle for any subset problem or its enumeration variant.

\paragraph{Fine-grained search space classification}
In our case, we assume the instance is represented as an undirected graph $G=(V,E)$.
Moreover, in contrast to previous approaches, we view \emph{individual vertices} of $G$ as classification problems as opposed to $G$ itself.
That is, the problem is to induce a mapping $\gamma : V \to \{0,1\} $ from a set of $L$ training examples $T = \{ \langle f(v_i), y_i \rangle \}^L_{i=1}$, where $v_i \in V$ is a vertex, $y_i \in \{0,1\}$ a class label, and $f : V \to \mathbb{R}^d$ a mapping from a vertex to a $d$-dimensional feature space.
We keep $d$ small and ensure that $f$ can be computed efficiently to ensure practicality.

\paragraph{Search strategies}
To learn the mapping $\gamma$ from $T$, we use a probabilistic classifier $P$ which outputs a probability distribution over $\{0,1\}$ for a given $f(u)$ for $u \in V$. 
We give two parameterized search strategies for enhancing a search algorithm $\mathcal{A}$ by $P$.
Define a \emph{confidence threshold} $q \in [0,1]$.
\begin{itemize}
\item \textit{Probabilistic preprocessing:} delete from $G$ each vertex predicted by $P$ to \emph{not} be in a solution with probability at least $q$, i.e., let $G' = G \setminus V'$, where $V' = \{ u \mid u \in V \wedge P(u = 0) \geq q \}$. 
Execute $\mathcal{A}$ with $G'$ as input instead of $G$.

\item \textit{Guiding by experience:}
Define a set of \emph{hints} $H = \{ u \mid u \in V \wedge P(u = 1) \geq q \}$ and use them to guide the search effort of $\mathcal{A}$ executed on input $G$.
\end{itemize}
The purpose of $q$ is to control the error and pruning rate of preprocessing: (i) it is more acceptable to not remove a vertex that is \emph{not} in a solution than to remove a vertex that \emph{is} in a solution, and (ii) a lower value of $q$ translates to a possibly higher pruning rate.

We observe that the probabilistic preprocessing strategy is a \textbf{heuristic}, i.e., it is possible the cost of an optimal solution in $G'$ differs from that in $G$.
However, the second strategy of guiding by experience is \textbf{exact}, i.e., given enough time, $\mathcal{A}$ will finish with a globally optimal solution.
It is also possible to combine the strategies by preprocessing first, and then executing $\mathcal{A}$ with $H$ defined on $G'$.
For the remainder of this work, we only focus on probabilistic preprocessing.

\section{Computational features}
\label{sect:features}
In this section, we describe the vertex features which can be computed efficiently and also capture fine-grained, localized structural properties of the graph.
Specifically, we employ the following features based on {\em graph measures} and {\em statistical properties}.

\paragraph{Graph measure based features} 
We use the following graph-theoretic features: \textbf{(F1)} number of vertices, \textbf{(F2)} number of edges, \textbf{(F3)} vertex degree, \textbf{(F4)} local clustering coefficient, and \textbf{(F5)} eigencentrality.

Features (F1), (F2), and (F3) capture the crude information regarding the graph, providing a reference for the classifier for generalizing to different distributions from which the graph might have been generated. 
Feature (F3), the \emph{degree} of $v$, denoted by $\deg(v)$, is the number 
of edges incident to~$v$.

Feature (F4), the {\em order-3 local clustering coefficient} (LCC) of a vertex 
is the fraction of its neighbors with which the vertex forms a triangle, encapsulating the {\em small world}~\cite{Watts1998} phenomenon.
In general, the \emph{order-$k$ local clustering coefficient} of $v$, denoted by $C_k(v)$, is computed as $C_k(v) = W_k(v) / {\deg(v) \choose k-1}$,
where $W_k(v)$ is the number of $(k-1)$-cliques in $G[N(v)]$, i.e., the subgraph of $G$ induced by the neighborhood of $v$ (excluding $v$ itself).
For computational efficiency, we limit our feature set to only order-three LCC.

Feature (F5), the {\em eigencentrality}, represents a high degree of connectivity of a vertex to other vertices, which in turn have high degrees as well. 
The \emph{eigenvector centrality} $\vec{v}$ is the eigenvector of the adjacency matrix $A$ of $G$ with the largest eigenvalue $\lambda$, i.e., it is the  solution of $\vec{A}\vec{v} = \lambda\vec{v}$.
The $i$th entry of $\vec{v}$ is the \emph{eigencentrality} of vertex $v$. In other words, this feature provides a measure of local ``denseness''. A vertex in a dense region shows higher probability of being part of a large clique.

\paragraph{Statistical features.} Intuitively, for a vertex $v$ present in a large clique, its degree and the local clustering coefficient would deviate 
from the underlying expected distribution characterizing the graph. 
Similarly, the neighbors of $v$ also present in the clique would demonstrate such a property. Interestingly, statistical features are not only computationally cheap but are also inherently robust in approximately capturing the local graph structural patterns as shown in~\cite{graph}. 

The above intuition is captured by the following features: \textbf{(F6)} the chi-squared value over vertex degree, \textbf{(F7)} average chi-squared value over neighbor degrees, \textbf{(F8)} chi-squared value over LCC, and \textbf{(F9)} average chi-squared value over neighbor LCCs.

Statistical significance can be captured by the notion of p-value~\cite{fitStatistics}. 
The {\em Pearson's chi-square statistic}, $\chi^2$, is a good estimate~\cite{pear} of the p-value, and for features (F6)-(F9) it is computed as
\begin{equation}
\label{eq:chis}
\chi^2 = \sum_{\forall i}\left[\left(O_i - E_i\right)^2 / E_i\right],
\end{equation}
where $O_i$ and $E_i$ are the observed and expected number of occurrences, of the possible outcomes $i$.

\paragraph{Classification framework}
To solve the problem described in Section~\ref{sec:framework}, we experiment with various (even non-probabilistic) classifiers via \texttt{auto-sklearn}~\cite{autosklearn} which is an automated system for algorithm selection and hyperparameter tuning (for a list of its classifiers, including non-linear ones, see~\cite[Table~1]{autosklearn}).
We observe highest accuracy with linear models, with negligible differences between logistic regression and (linear) support vector machines.

Thus, we use a linear classifier (logistic regression) trained for~400 epochs with stochastic gradient descent.
We use a standard L2 regularizer, and use 0.0001 as the regularization term multiplier determined by a grid search.
We use a standard implementation (SGDClassifier) from \texttt{scikit-learn}~\cite{scikit-learn}.

\section{Experimental results}
\label{sect:experimental}
In this section, we demonstrate that supervised learning is viable for solving large structured instance of $\NP$-hard problems, whereas previous approaches relying have only scaled up to instances with 200 vertices.
Furthermore, as mentioned in~\cite{Eblen2012}, it is well-known that the topology of real-world networks differ from that of random graphs.
Hence, we show scalability to large real-world networks.

\begin{table*}[t]
\caption{Pruning ratios and speedups.
The column ``\texttt{fmc} ($\omega$)'' is the maximum clique size found by the \texttt{fmc} heuristic, followed by the true maximum clique size $\omega$ in parenthesis.
The column $P_d$ is the pruning ratio obtained by the degree method, while $P_o$ is the pruning ratio of our framework.
The columns ``\texttt{cliquer}(d)'' and ``\texttt{cliquer}(o)'' give the runtimes of \texttt{cliquer} on \emph{only} degree-pruned instances and on instances pruned by our framework, respectively.
The confidence threshold $q=0.55$, all results are averaged over 3 runs and include feature computation time, and the clique accuracy remains one.
In the last four columns, the parenthesis show the speedup while an out of memory for the original instance is marked with *.
}
\label{tbl:big_graphs}
\centering
\small
\def\arraystretch{0.5}
\setlength{\tabcolsep}{5pt}
\begin{tabular}{llllllllll}
\toprule 
\textbf{Instance} & $|V|$ & $|E|$ & \texttt{fmc} ($\omega$) & $P_d$ & $P_o$ & \textbf{igraph} & \textbf{EmMCE} & \textbf{cliquer}(d) & \textbf{cliquer}(o) \\ 
\midrule
socfb-A-anon 			& 3 M & 24 M & 23 (25) & 0.85 & 0.94 & 101.97 (20.84) & 113.90 \textbf{(44.82)} & 1337.85 (*) & 211.06 (*) \\
\rowcol socfb-B-anon 	& 3 M & 21 M & 23 (24) & 0.86 & 0.94 & 86.64 (22.26) & 96.34 (41.96) & 1038.59 (*) & 193.50 (*) \\
socfb-Texas84			& 36 K & 2 M & 44 (51) & 0.37 & 0.76 & 56.88 (1.29) & 48.62 (1.28) & 4.10 (2.19) & 1.87 \textbf{(4.79)} \\
\midrule
bio-WormNet-v3			& 16 K & 763 K & 90 (121) & 0.68 & 0.90 & 4400.12 (1.22) & 4593.47 (1.26) & 1.04 (6.69) & 0.89 \textbf{(7.87)} \\
\midrule

\rowcol web-wikipedia2009 	& 2 M & 5 M & 31 (31) & 0.98 & 0.99 & 1.50 (412.79) & 1.98 (1574.39) & 8.86 (*) & 3.96 (*) \\
web-google-dir 				& 876 K & 5 M & 44 (44) & 0.97 & 0.98 & 0.39 (150.20) & 0.67 (991.52) & 4.40 (*) & 2.07 (*) \\

\midrule

soc-flixster  			& 3 M & 8 M & 29 (31) & 0.97 & 0.99 & 23.71 \textbf{(54.66)} & 18.64 (20.25) & 44.27 (*) & 9.52 (*) \\
\rowcol soc-google-plus & 211 K & 2 M & 56 (66) & 0.94 & 0.97 & 6620.77 (1.09) & 6711.56 (1.32) & 1.50 (214.20) & 1.04 \textbf{(310.40)} \\
soc-lastfm 				& 1 M & 5 M & 14 (14) & 0.92 & 0.97 & 9.47 (16.79) & 9.12 (31.23) & 65.04 (*) & 11.61 (*) \\
\bottomrule 
\end{tabular}
\end{table*}

\subsection{Exact clique-finding algorithms}
We use the following state-of-the-art algorithms:
\begin{itemize}
\item \texttt{igraph}~\cite{igraph} software implementation of a modified Bron-Kerbosch algorithm~\cite{Eppstein2010}.

\item \texttt{EmMCE}~\cite{Cheng2011}, an external memory algorithm focusing on low I/O complexity.

\item \texttt{cliquer}~\cite{Niskanen2003}, using the branch-and-bound algorithm of~\cite{Ostergard2002}.

\end{itemize}
Note that all algorithms here are \emph{exact}, i.e., they will indeed return all optimal solutions.
Strictly speaking, also note that \texttt{igraph} and \texttt{EmMCE} solve the more general problem of \emph{maximal} clique enumeration.

We also experimented with the \texttt{MoMC} solver of~\cite{Li2017}, but it reported out of memory for large networks. Runs on this solver are thus omitted.
The experiments for real-world networks (Subsection~\ref{subs:real}) are executed on an Intel Core i7-4770K CPU (3.5 GHz), 8 GB of RAM, running Ubuntu 16.04.
As an exception, we run experiments for \texttt{cliquer} on Intel Xeon E5-2680 and 102 GB of RAM.

\subsection{Real-world networks}
\label{subs:real}
To demonstrate the practicality and usefulness of our framework, we use real-world networks from Network Repository~\cite{Rossi2015} (\url{http://networkrepository.com/}).
In particular, we consider biological networks, Facebook networks, social networks, and web networks, which contain 36, 114, 57, and 27 networks, respectively.
For training, we choose from the biological networks 32 smallest networks, from the Facebook networks 109 smallest networks, from the social networks a small sample of 32 networks, and from the web networks a small sample of 11 networks.
For reasons of space, we omit the exact details of the training networks.
For testing, we use from each domain some of the largest networks by edge count, detailed in Table~\ref{tbl:big_graphs}.

\paragraph{Setup and accuracy measures}
We implement the probabilistic preprocessing strategy of Section~\ref{sec:framework} with the classifier $P$ as described in Section~\ref{sect:features}.
We fix the confidence threshold $q = 0.55$.
We consider the following accuracy measures.
The \emph{pruning ratio} is the ratio of the number of vertices predicted by $P$ to not be in a solution with probability at least~$q$ to the number of vertices $n$ in the original instance.
The \emph{clique accuracy} is one iff the number of \emph{all} maximum cliques of the instance $G$ is equal to the number of all maximum cliques of the reduced instance $G'$ and $\omega(G) = \omega(G')$, where $\omega(G)$ is the size of a maximum clique in $G$.

\paragraph{Preprocessing and comparison}
A safe preprocessing strategy is the \emph{degree method}: find a clique of size $k$, and delete every vertex with degree less than~$k - 1$.
We implement this by using Fast Max-Clique Finder~\cite{Pattabiraman2013} (\texttt{fmc}), a state-of-the-art heuristic for finding a maximum clique, designed for real-world networks, typically finding near-optimal solutions (see Table~\ref{tbl:big_graphs}).
The \texttt{fmc} heuristic runs in less than 2 seconds for each network.

We make use of the degree method in two ways.
First, being a well-known preprocessing method, we compare our method against it.
Second, the degree distributions of some real-world networks follow a power law.
That is, such networks have many low degree vertices.
In such cases it might not be surprising that an algorithm learns to remove many vertices.
Thus, to avoid such a situation from impacting our results, we preprocess each training and test instance first using the degree method.
This helps our framework to discard non-trivial vertices, going beyond the degree method.

\paragraph{Features, classifier, and vertex classification accuracy}
We train four classifiers that we name \texttt{bio31}, \texttt{socfb107}, \texttt{soc32}, and \texttt{web13} using the mentioned networks.
More precisely, for each network $G_i$ in the training set of the three domains, we list \emph{all} maximum cliques $\mathcal{C}_i = \{ C_1, C_2, \ldots, C_\ell \}$ in $G_i$, use as label-0 examples the vertices in $H = \bigcup_{i=1}^{\ell} V(C_i)$, and to create a balanced dataset, sample uniformly at random $|H|$ vertices from $V(G_i) \setminus H$ and use them as label-1 examples.
The final training set is obtained by computing the feature vectors of each vertex.
We use the features described in Section~\ref{sect:features}.
In particular, \texttt{bio31} is trained with 2178 feature vectors, \texttt{socfb107} with 10746 feature vectors, \texttt{soc32} with 2008 feature vectors, and \texttt{web11} with 2556 feature vectors.
A 4-fold cross validation over the 2178 feature vectors gives an average \textit{vertex classification accuracy} of~0.96, the same over the 10746 feature vectors results in an average of~0.93, the same over the 2008 feature vectors results in an average of~0.97, and the same over the 2556 feature vectors results in an average of~0.80.

We implement feature computation in C++, taking less than 12 seconds for each test network.
Further optimization is also possible in terms of e.g., parallelization.

\paragraph{Pruning ratios and speedups}
We show the pruning ratios and solve times along with solver speedups in Table~\ref{tbl:big_graphs}.
For each graph, \textbf{we retain all optimal solutions} (i.e., the clique accuracy is one), and obtain \textbf{speedups as large as 300x}, even for the branch-and-bound strategy of \texttt{cliquer}.
It can also be seen that our strategy speeds up \texttt{cliquer} by around 6x more than the degree method alone (compare two last columns in Table~\ref{tbl:big_graphs}).
Also importantly, our framework safely prunes away around 90 \% or more of the vertices, considerably shrinking the input sizes which is a fundamental issue in applications in e.g., computational biology~\cite{Eblen2012}.
This alleviates memory issues with state-of-the-art algorithms as reported in~\cite{Cheng2011}.

We stress that our approach works \textbf{without any knowledge or dependence on an estimate of $\omega$} at runtime, while the quality of the degree method crucially depends on it.




\section{On supervised learning for hard problems}
\label{sec:planted}
The goal of this section is two-fold: 
(i) to explain the high accuracy of our proposed framework, even when it was trained with small instances, and
(ii) as a consequence, argue that supervised learning is a viable approach for solving structured instances of certain hard problems.

To ensure that the input instances are, at some point, ``structure-free'' we turn to the following heavily-studied variant of the maximum clique problem.
This serves as a representative of the \emph{worst-case input} for our preprocessing strategy.
Also, observe that in case the input graph has a unique maximum clique, MCE is equivalent to finding the (single) maximum clique.

\begin{figure}[t]
\centering
	\begin{tabular}{ccc}
	\hspace*{-4mm}	\includegraphics[width=0.33\columnwidth]{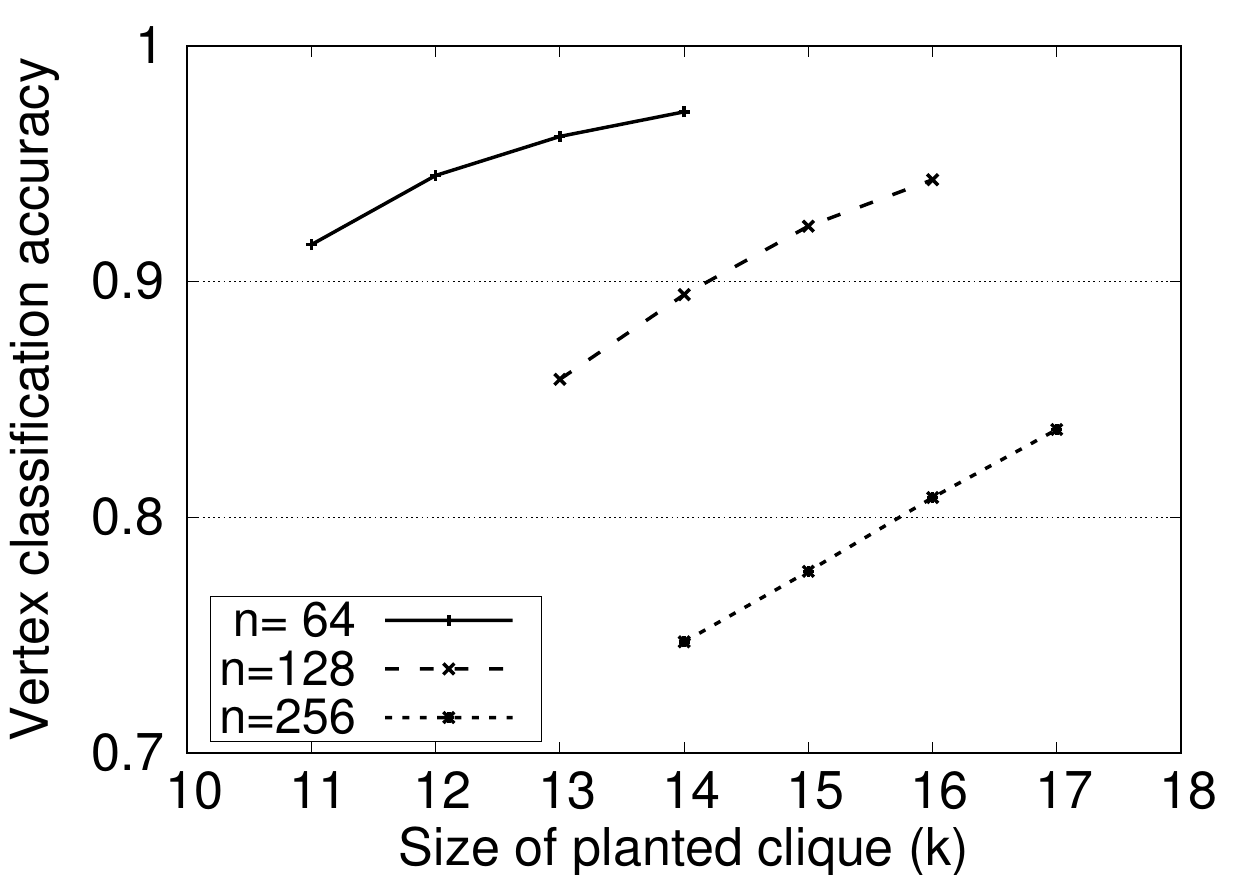} &
	\hspace*{-6mm}	\includegraphics[width=0.33\columnwidth]{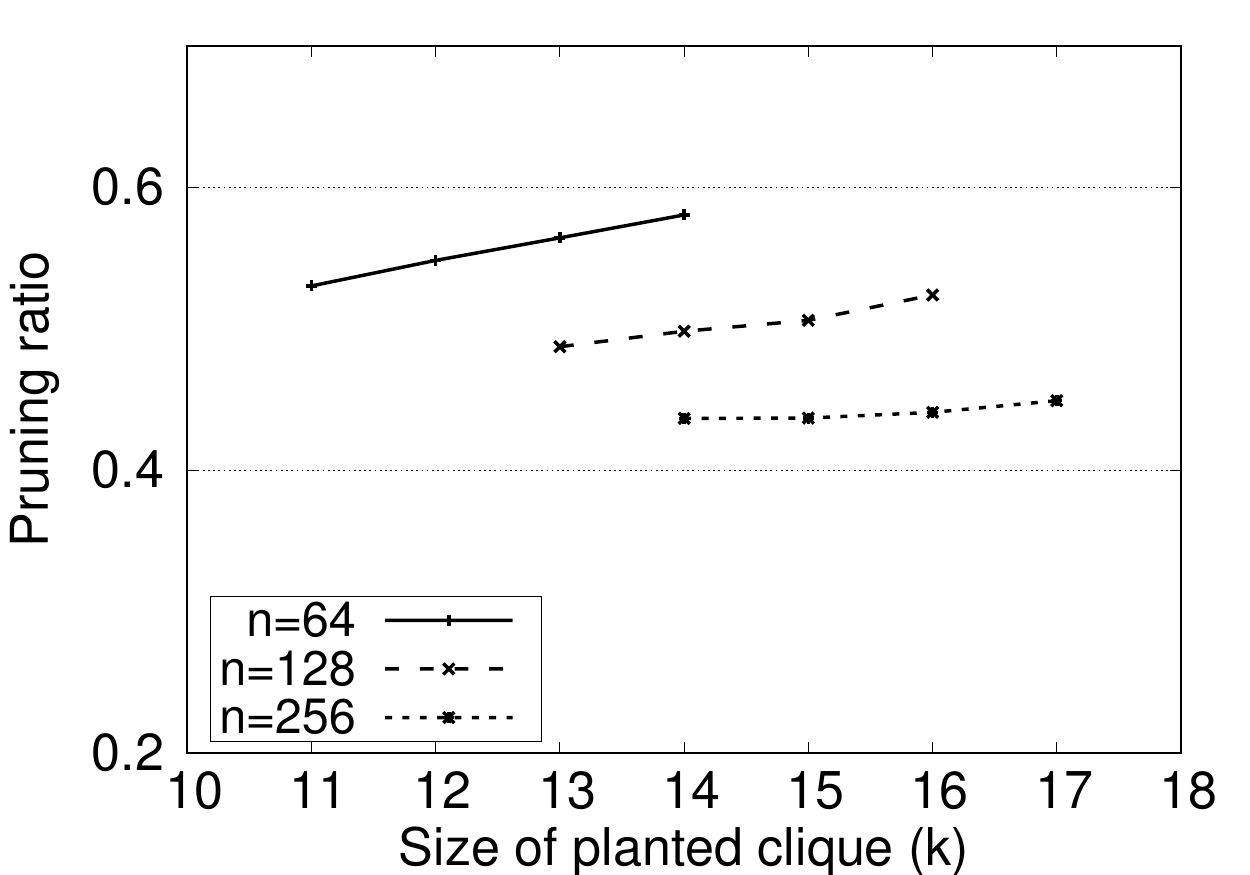} &
	\hspace*{-6mm}	\includegraphics[width=0.33\columnwidth]{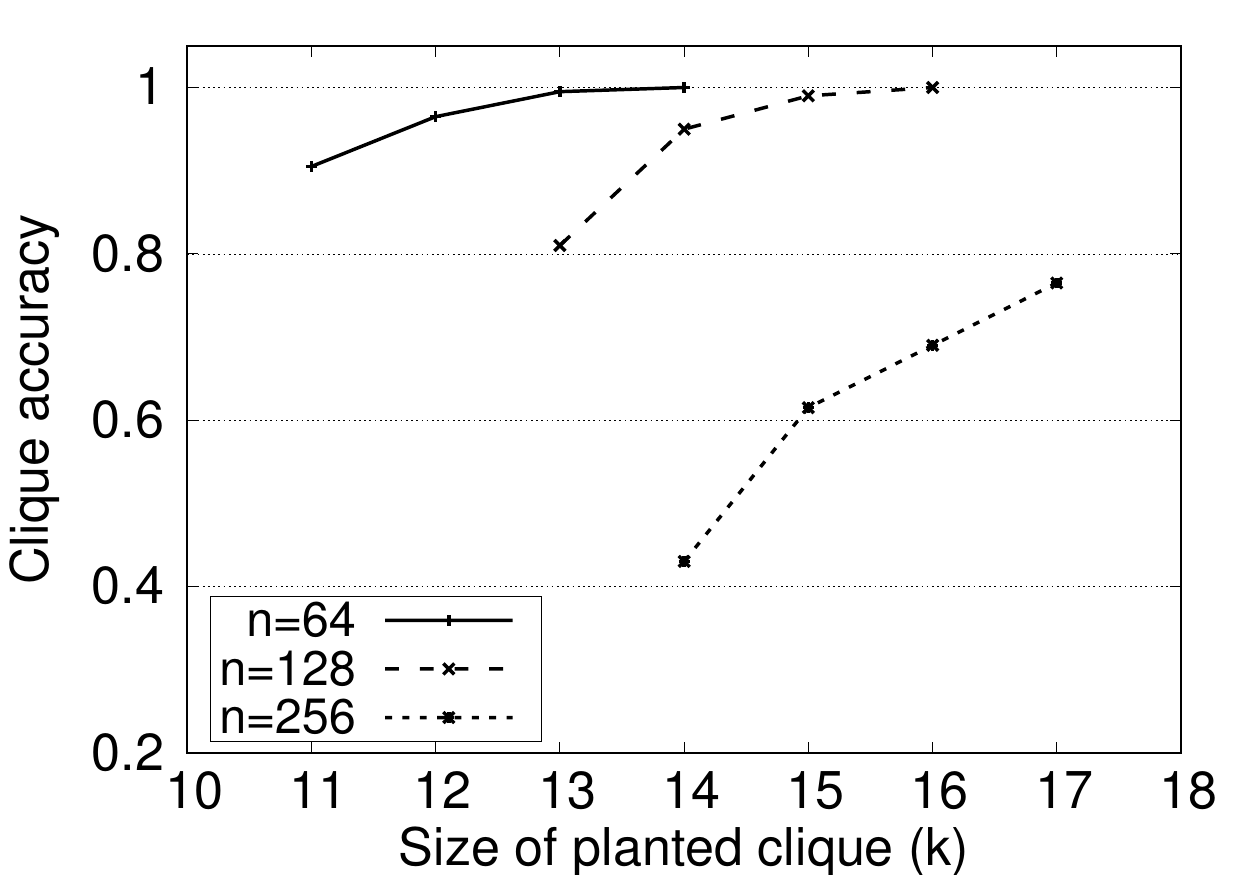} \\
	(a) Vertex acc. & (b) Pruning ratio & (c) Clique acc.
	\end{tabular}
	\caption{The vertex accuracy, pruning ratio, and clique accuracy of our framework when trained with $G(n,1/2)$ with three different parameter pairs $(64,10)$, $(128,12)$, and $(256,13)$. The predictions are for independent, distinct samples with the same $n$, but growing planted clique size~$k$.}
	\label{fig:acc}
\end{figure}

\subsection{Planted clique}
\label{subs:plantedc}
In the \emph{planted clique problem}~\cite{Jerrum1992,Kucera1995}, we are given an Erd\H{o}s-R\'{e}nyi random graph~\cite{Erdos1959} $H := G(n,p)$, i.e., an $n$-vertex graph where the presence of each edge is determined independently with probability $p$.
In addition, the problem is parameterized by an integer $k$ such that a random subset of $k$ vertices has been chosen from $H$ and a clique added on it.
On this input, the task is to identify (with the knowledge of the value of $k$) the $k$ vertices containing the planted clique.

The problem is easy for $k \leq \log_2(n)$.
In particular, as shown in~\cite{Bollobas2013}, the clique number of $G(n,p)$ as $n \to \infty$ is almost surely $w$ or $w+1$ where $w$ is the greatest natural number such that
\begin{equation}
\label{eq:clnum}
{n \choose w} p^{w \choose 2} \geq \log(n),
\end{equation}
where $w$ is roughly $2 \log_2 (n)$.
Even when a clique of such size is known to exist (whp), we only know how to find a clique of size $\log_2(n)$ efficiently,\footnote{It is conjectured~\cite{Karp1976,Feldman2017} that there is no polynomial-time algorithm for finding a clique of size $(1+\eps) \log_2 n$ for any $\eps > 0$ in $G(n,1/2)$.} and also solve the problem in polynomial-time when $k$ is large enough.
Specifically, it is known that several algorithmic techniques such as spectral methods (see e.g.,~\cite{Feldman2017} for more) produce efficient algorithms for the problem when $k = \Omega(\sqrt{n})$.

However, settling the complexity of the problem is a notorious open problem when $k$ is between $2 \log_2(n)$ and $\sqrt{n}$. Next, we will focus precisely on this difficult region.

\subsection{Pushing the limits of preprocessing}
\label{subs:limits}
In this subsection, we explore the limits of scalability and robustness of our framework on the planted clique problem.
All experiments are done on an Intel Core i5-6300U CPU (2.4 GHz), 8 GB of RAM, running Ubuntu 16.04, differing only slightly from the earlier hardware configuration.
For all experiments here, we use only the \texttt{igraph} algorithm.

\paragraph{Generation of synthetic data}
We use the \texttt{genrang} utility program~\cite{McKay2014} to sample a random graph~$H := G(n,p)$.
To plant a clique of size $k$, we sample uniformly at random~$k$ vertices, denoted by $K$, from~$H$ and insert all corresponding at most ${k \choose 2}$ missing edges into~$H$.

For each $H$, we compute the features described in Section~\ref{sect:features} with the following differences: we add (F10) the order-four LCC and modify (F8) and (F9) to consider order-four LCC instead of the LCC.
As explained in Section~\ref{sect:features}, this brings more predictive power while still remaining computationally feasible for small graphs.
The values $E_i$ in Equation~\ref{eq:chis} for (F6) and (F7) are the expected degree $n \cdot p$, while for (F8) and (F9) they are the expected order-$k$ LCC given as ${n - 1 \choose k-1} p^{k \choose 2} \cdot {1} / {np \choose k-1}$. 
To ensure a balanced dataset, we sample (i) $k$ label-0 examples from $K$ and (ii) $k$ label-1 examples from $G \setminus K$, both uniformly at random.

For training, we consider $n \in \{64,128,256,512\}$ because the clique number grows roughly logarithmically with $n$ (see Equation~\ref{eq:clnum}).
We fix $p = 1/2$.
For every $n$, we compute $w$ from Equation~\ref{eq:clnum}, and sample graphs $G(n,p)$ with a planted clique of size $k \in \{w+2,\ldots,w+6\}$ such that each pair $(n,k)$ gives a dataset of size at least 100K feature vectors.
When planting a clique of size at least $w+2$, we try to guarantee the existence of a unique maximum clique in the graph.
However, this procedure does not always succeed due to randomness, but we do not discard such rare outcomes. 

\begin{figure}[t]
\centering
	\begin{tabular}{ccc}
	\includegraphics[width=0.28\columnwidth]{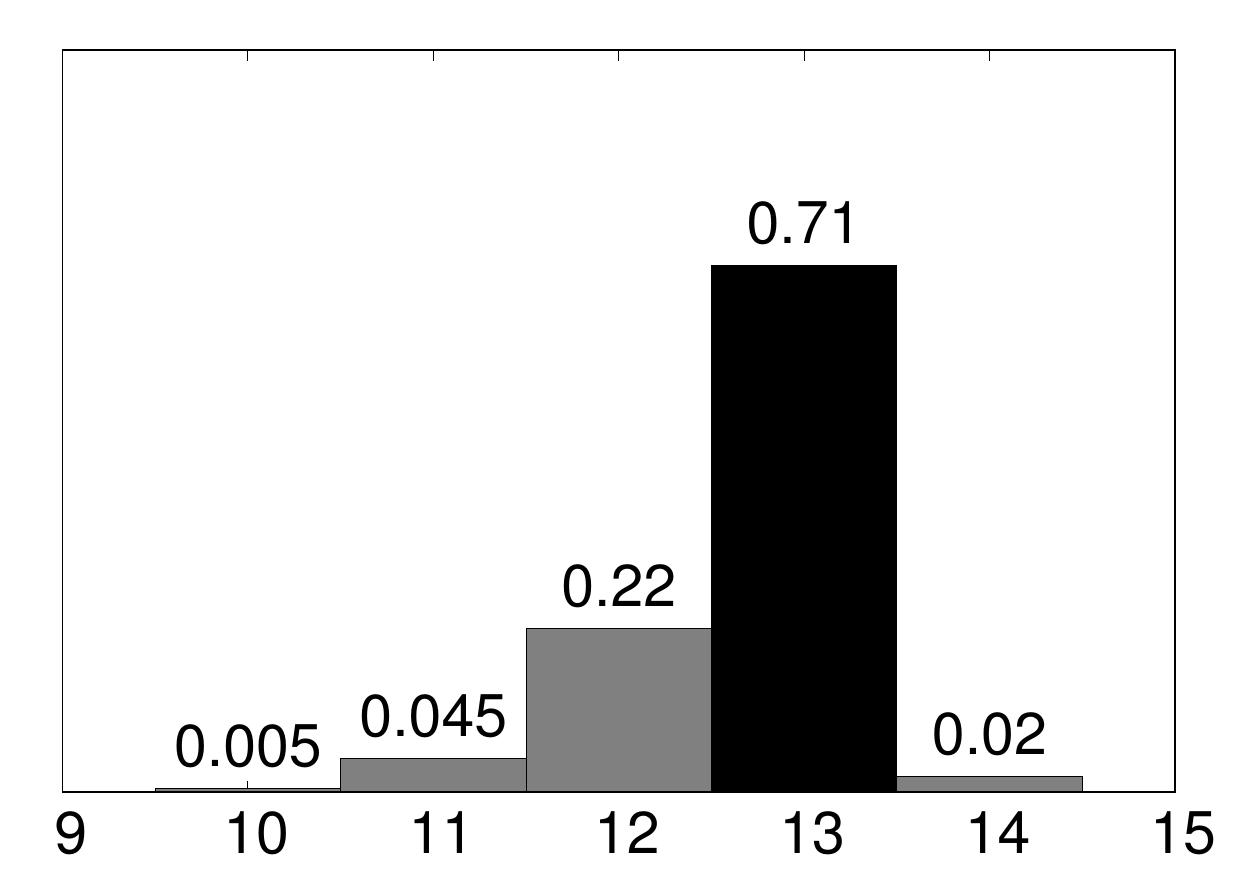} &
	\includegraphics[width=0.28\columnwidth]{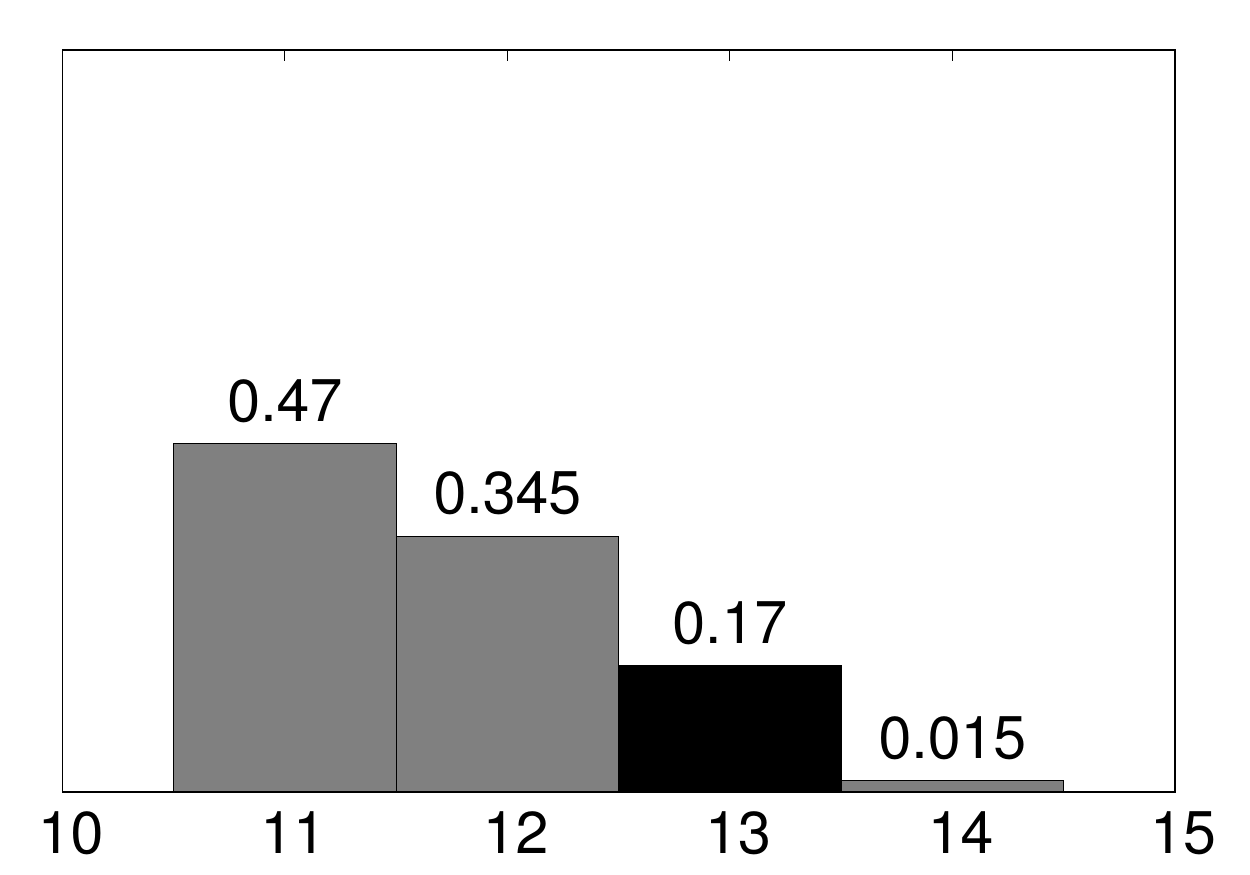} &
	\includegraphics[width=0.28\columnwidth]{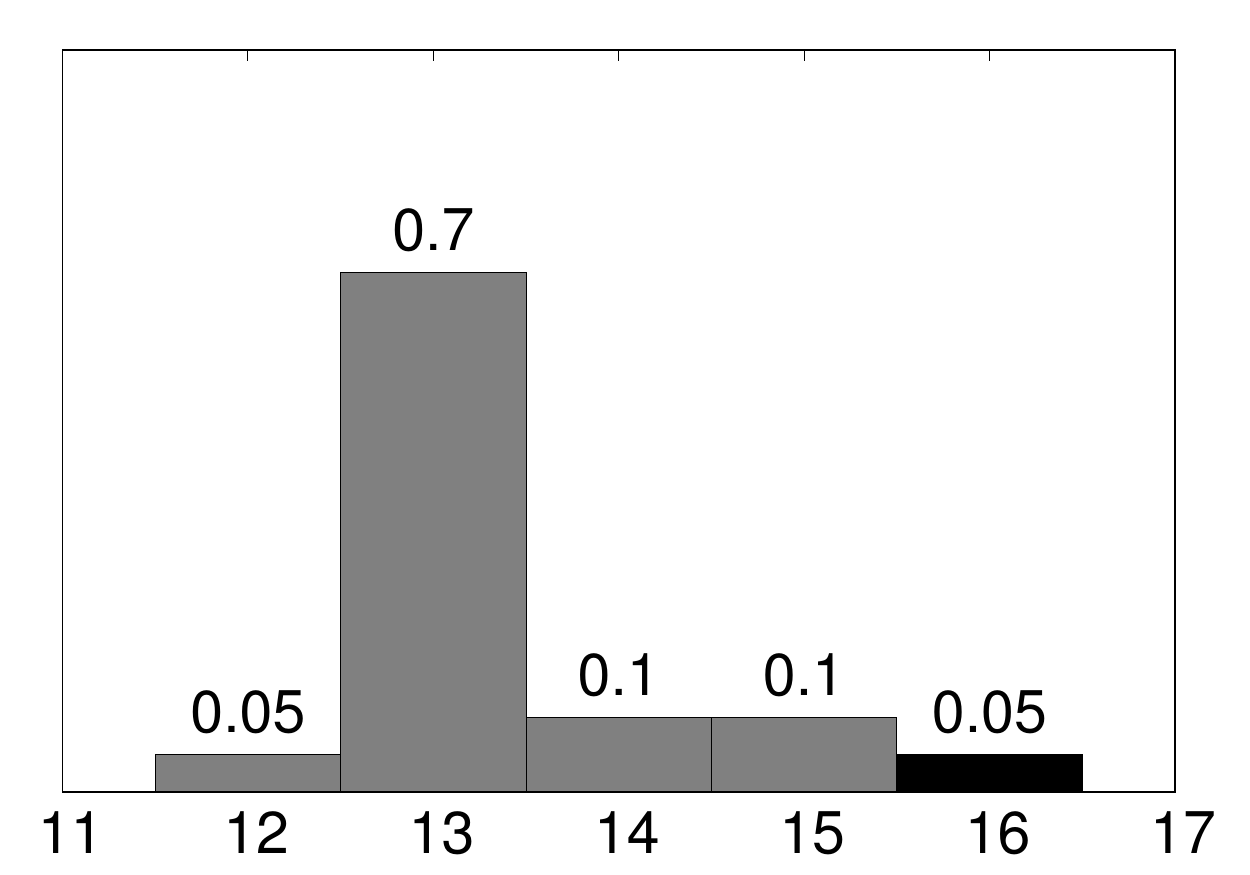} \\
	(a) $(128,13)$ & (b) $(256, 13)$ & (c) $(512, 16)$
	\end{tabular}
	\caption{Distribution of extracted maximum clique size, with black bars denoting the size of the planted clique. Both (a) and (b) are over 200 samples, while (c) is over 20 samples. In each, the predicting classifier has been trained with 64-vertex random graphs with a planted clique of size 10.}
	\label{fig:cldist}
\end{figure}

\begin{table*}
  \centering
  \small
  \caption{Robustness and speedups with fixed $n$ and increasing $k$. The leftmost two columns show the data $(n,k)$ used to train a classifier $P$. For each planted clique size $k+1$, $k+2$, and $k+3$, we show the average pruning ratio (column ``Pruned''), the average clique accuracy (column ``Acc.'), the average runtime of \texttt{igraph} on the reduced instance obtained from our framework using $P$ (column ``Time~(s)''), and the average speedup over executing the same algorithm on the original instance.}
  \label{tbl:robust}
  \def\arraystretch{1}
  \setlength{\tabcolsep}{4pt}
  \begin{tabular}{*{111}{l}}
    \toprule
    & & & \multicolumn{2}{c}{$k+1$} & & & \multicolumn{2}{c}{$k+2$} & & & \multicolumn{2}{c}{$k+3$} \\
    \cmidrule(lr){3-6}
    \cmidrule(lr){7-10}
    \cmidrule(lr){11-14}
    $n$ & $k$ & Pruned & Acc. & Time (s) & Speedup & Pruned & Acc. & Time (s) & Speedup & Pruned & Acc. & Time (s) & Speedup \\
    \midrule
    \phantom{0}64 & 10 & 0.530 & 0.905 & 0.068 & 0.132 & 0.548 & 0.965 & 0.068 & 0.135 & 0.564 & 0.995 & 0.068 & 0.135 \\
    \rowcol 128 & 12 & 0.506 & 0.710 & 0.301 & 0.759 & 0.517 & 0.875 & 0.296 & 0.774 & 0.525 & 0.935 & 0.297 & 0.784 \\
    256 & 13 & 0.489 & 0.170 & 3.261 & 3.264 & 0.493 & 0.190 & 3.233 & 3.304 & 0.493 & 0.310 & 3.260 & 3.315 \\
    \rowcol 512 & 15 & 0.492 & 0.05 & 70.587 & 12.994 & 0.492 & 0.05 & 70.086 & 12.816 & 0.491 & 0.100 & 70.562 & 12.722 \\
    \bottomrule
  \end{tabular}
\end{table*}

\paragraph{Vertex classification accuracy}
We study the accuracy of our classifiers for distinguishing vertices that are and are not in a maximum clique.
Specifically, we train a classifier for each pair $(n,k) \in \{ (64,10), (128, 12), (256, 12) \}$, and test for unseen graphs with the same $n$ but growing planted clique size $k' = k + 1, \ldots, k+4$.
The results are shown in Figure~\ref{fig:acc}~(a).
As expected, the classification task becomes easier once $k'$ increases.
This is also supported the fact that multiple algorithms solve the planted clique problem in polynomial-time for large enough $k'$ (see Subsection~\ref{subs:plantedc}).
In addition, as $n$ grows larger, we see accuracy deterioration caused by the converge of the local properties towards their expected values.
Especially for small values of $k'$, the injection of the planted clique is not substantial enough to cause significant deviations from the expected values.

\paragraph{Pruning ratio and clique accuracy}
We study the effectiveness of our framework as a probabilistic preprocessor for the planted clique instances.
We fix the confidence threshold $q = 0.55$ and use the same set of classifiers and test data.
The average pruning ratios over all instances are shown in Figure~\ref{fig:acc}~(b).
We see pruning ratios as high as at most~0.6, while always discarding more than 40~\% of the vertices.

Now, it is possible that $P$ makes an erroneous prediction causing the deletion of a vertex, which in turn lowers the size of a maximum clique in the instance (although recall that this was \emph{not} observed in Section~\ref{sect:experimental} for real-world networks).
The average clique accuracies over all instances are shown in Figure~\ref{fig:acc}~(c).
Here, we see that for $n=256$, the vertex accuracy (Figure~\ref{fig:acc}~(a)) is still above 0.7, but the clique accuracy drops to above 0.4.
As the vertex accuracy decreases, the probability of deleting a vertex present in a maximum clique increases, translating to a higher chance of error in extracting a maximum clique.
However, while not completely error-free, we observe that even in the case of $(256,13)$ we always delete at most two members of a maximum clique, whereas in the case of $(512,16)$, 95 \% of the time, we extract a maximum clique of size at least~13 (see Figure~\ref{fig:cldist}).

\paragraph{Robustness and speedups}
The robustness and speedups obtained using the \texttt{igraph} algorithm are given in Table~\ref{tbl:robust}.
Here, the clique accuracy and runtime are obtained as the average over 200 samples for each $(n,k)$ except for $(n,k) = (512,15)$ for which there are 20 independent samples.
We see the drop in clique accuracy when a classifier $P$ is trained with $(n,k) \in \{ (256,12), (512,15) \}$ and is predicting for the same $n$ but increasing~$k$.
The clique accuracy is a strict measure, so to quantify the severeness of the erroneous predictions made by $P$ we show the distributions of the extracted maximum clique sizes in Figure~\ref{fig:cldist} for some pairs $(n,k)$.
Again, we observe the effects of growing $n$ causing the convergence of local properties, consequently decreasing the predictive power of $P$.
For $(n,k) = (128,13)$, 73 \% of the runs still produce an optimal solution (here, one can also observe the rare event of having a maximum clique of size~14 when the planted clique was of size~13).

\paragraph{The case for supervised learning on intractable problems}

\begin{table}[t]
  \centering
  \small
  \caption{Deviation in vertex classification accuracy.}
  \label{tbl:retrain}
  \def\arraystretch{1}
  \setlength{\tabcolsep}{4pt}
  \begin{tabular}{llcc}
    \toprule
    $n$ & $k$ & Trained acc.\ & Rob.\ acc.\ \\
    \midrule
    128 & 12 & 0.858 & 0.844 \\
    \rowcol 256 & 13 & 0.747 & 0.728 \\
    512 & 15 & 0.678 & 0.665 \\
    \bottomrule
  \end{tabular}
\end{table}

As $n$ grows, the instances get increasingly time-consuming to solve even for state-of-the-art solvers for suitable $k$, as there is no exploitable structure.
Consequently, obtaining optimally labeled data becomes practically impossible for large enough $n$.
However, up to a point, random graphs with $n = 64$ and $k = 10$ are representative of the input for large graphs as well, and obtaining the optimal label for such a small graph is fast.

{We show the deviation in vertex classification accuracy in Table~\ref{tbl:retrain}.}
The column ``Trained acc.'' corresponds to the accuracy of the classifier trained with the values $n$ and $k$ mentioned in the two first columns, while the column ``Rob.\ acc.'' is the accuracy of a classifier trained with smaller $(n,k) = (64,10)$ instances, and predictions are made for the specified $(n,k)$ with planted clique size $k + 1$.
A key observation is that the difference between the two accuracies in a single row in Table~\ref{tbl:retrain} is small enough not to warrant training on larger instances.
This offers an explanation for the perfect clique accuracy with limited training, as observed in Section~\ref{sect:experimental} for real-world networks.
This observation reduces the need of labeling costly data points for training.

\section{Discussion and conclusions}
We proposed a simple, powerful, and flexible machine learning framework for (i) reducing the search space of computationally difficult enumeration problems and (ii) augmenting existing state-of-the-art solvers with informative cues arising from the input distribution.
In particular, we focused on a probabilistic preprocessing strategy, which retained \emph{all} maximum cliques on a representative selection of large real-world networks from different domains. 
We showed the practicality of our framework by showing it speeds up the execution of state-of-the-art solvers on large graphs without compromising the solution quality.
In addition, we demonstrated that supervised learning is a viable approach for tackling $\NP$-hard problems in practice.

For future work, we will perform more extensive experiments on a wider set of instances, and provide a deeper analysis of our model.

\begin{quote}
\begin{small}
\bibliographystyle{aaai}
\bibliography{bibliography}
\end{small}
\end{quote}

\end{document}